\documentclass[conference]{IEEEtran}
\IEEEoverridecommandlockouts
\usepackage{cite}
\usepackage{amsmath,amssymb,amsfonts}
\usepackage{algorithmic}
\usepackage{graphicx}
\usepackage{textcomp}
\usepackage{xcolor}
\usepackage{times}
\usepackage{epsfig}
\usepackage{graphicx}
\usepackage{amsmath}
\usepackage{amssymb}
\usepackage{booktabs}

\usepackage{cite}
\usepackage{amsmath,amssymb,amsfonts}
\usepackage{algorithmic}
\usepackage{graphicx}
\usepackage{textcomp}
\usepackage{xcolor}
\usepackage{ulem}
\usepackage{comment}
\usepackage{caption}
\usepackage{subcaption}
\usepackage{multirow}
\usepackage{multirow}
\usepackage{tabularx}
\usepackage{array}
\usepackage{float}
\usepackage[papersize={8.5in,11in},top=0.75in, bottom=1in, right=0.68in, left=0.68in]{geometry}

\def\BibTeX{{\rm B\kern-.05em{\sc i\kern-.025em b}\kern-.08em
    T\kern-.1667em\lower.7ex\hbox{E}\kern-.125emX}}
\begin{document}

\title{Obscured Wildfire Flame Detection by Spatio-temporal Analysis of Smoke Patterns Using Frame-wise Transformers\\
}

\author{\IEEEauthorblockN{1\textsuperscript{st} Uma Meleti}
\IEEEauthorblockA{\textit{School Of Computing} \\
\textit{Clemson University)}\\
Clemson, USA \\
umeleti@clemson.edu}
\and
\IEEEauthorblockN{2\textsuperscript{nd} Abolfazl Razi}
\IEEEauthorblockA{\textit{School Of Computing} \\
\textit{Clemson University}\\
Clemson, USA \\
arazi@clemson.edu}
\and
\IEEEauthorblockN{3\textsuperscript{rd} Fatemeh Afghah}
\IEEEauthorblockA{\textit{Department of Electrical Engineering} \\
\textit{Clemson University}\\
Clemson, USA \\
fafghah@clemson.edu}
}

\maketitle

\begin{abstract}
    In this paper, we propose a new paradigm in wildfire management for detecting both visible and obscured fire using Deep Learning (DL). 
    The essence of our method is the joint spatial and temporal analysis of RGB videos, embracing the fact that temporal patterns of smoke motions in consecutive video frames can indicate obscured fire flames. This perspective contrasts the traditional approach of independently analyzing images and individual video frames.
    To this end, we propose a frame-wise transformer architecture that applies an attention mechanism between the consecutive frames while preserving spatial information within frames as opposed to regular attention mechanisms like the non-local block,
    which combines spatial and temporal information into a single vector. We also optimize the temporal pattern lengths under different spatial resolutions to achieve the highest performance while reducing the computational cost. 
    To investigate the applicability of our method in aerial image processing, we applied it to a curated version of the FLAME2 dataset that includes dual-mode RGB/IR videos. The IR video feed is used to extract ground truth, while the model is trained only on RGB videos to maintain applicability to low-cost commercial drones (which are equipped only with regular cameras) and facilitate processing existing RGB-only datasets.     
    Our proposed method achieves a Segmentation Foreground Dice score of 92.61\%, an Object Detection Precision and Recall rate of 93.21\% and 91.73\%, and clip-level classification accuracy of 98.02\%, respectively. We also investigated the efficacy and generalizability of our method through ablation study \footnote{This material is based upon work supported by the National Science Foundation under Grant Numbers CNS2204721 and CNS-2204445.}.
    
    
   %
\end{abstract}



\section{Introduction}
Wildfires have become prevalent and destructive in many parts of the world. Regardless of the cause, wildfires can have severe consequences, including the loss of human lives, destruction of property, disruption of wildlife, food production, and crop supply chain as well as significant environmental damage.  Once a wildfire starts, there are various ways to monitor and control these wildfires, including observation towers, direct human intervention, satellite imaging, and manned aircraft. Using drones is one of the most efficient ways of fire monitoring due to its low operation cost, customizable sensing and imaging, flexible operation, and ease of use in harsh environments by advanced flight control features and partial autonomy (e.g., safe auto-landing). 

    One of the most significant advantages of using drones in wildfire fighting is the ability of gathering real-time data about the fire’s behavior. Equipped with cameras and other sensors, drones can fly over the fire and capture relevant information. Drones can also be equipped with water tanks or fire extinguishers containing fire-retardant chemicals, such as carbon dioxide (CO2), potassium bicarbonate (KHCO3), or evaporating like bromochlorodifluoromethane (CF2ClBr) to be dropped on hot spots to create fire breaks. The efficient utilization of drones will significantly advance controlling these fires. 

    Targeting places where actual burning is happening, instead of blindly spraying fire retardant gases everywhere, will help set these fires off quickly and efficiently. Finding out these burning places in real-time is very difficult, especially when fire flames are obscured by thick smoke. An infrared camera can help find these hidden fires, but IR cameras are expensive when utilized on a large scale. 

    We devised a methodology that uses an RGB camera feed, analyzes the video frames sequentially, and detects the obscured fires using temporal features, such as smoke patterns. We show that such features can be indicative of the fire's exact location and temporal behavior. We formulated the problem as a semantic segmentation of the obscured fire by analyzing the sequence of frames in the video. 

    We propose a deep learning architecture, ‘Frame-Wise Transformer,’ that utilizes a pre-trained CNN backbone to extract the features from video frames and passes through a convolutional transformer that aggregates the video features and sends them to a decoder, which will predict the segmentation map of the fire flames. We have implemented a convolutional transformer mimicking the NLP transformer, treating the video frame features as word embedding. Utilizing the popular non-local block \cite{DBLP:journals/corr/abs-1711-07971} for attention mechanism did not yield reasonable results. Therefore, we re-formulated the attention mechanism to restrict attention only between frames while not mixing within the frame information.  
    We modified the non-local block accordingly and achieved a significant performance gain. To balance the spatial resolution and temporal depth, we examined different frame rates while aggregating features at different resolutions. We observed that using more frames at lower resolutions provides higher performance while reducing the computation cost compared to using equal depth at all resolutions. The overall design can be extended to other video-based deep-learning tasks by making minor modifications to accommodate task-specific requirements.

    To this end, we curated a dataset from the existing FLAME2 \cite{shamsoshoara2021aerial} dataset by selecting part of video frames from the original data where the drone is stationary, and there is high synchronization with the IR images to avoid misalignment errors. We also use the corresponding IR videos to extract ground truth for our task by performing a series of image-processing operations. We have visually verified the correspondence of RGB video with the processed IR video for frame-level synchronization. 

    The rest of the paper is organized as follows. We highlight the unique features of our method compared to previous methods in Section 2, then proceed by elaborating on the details of the generated dataset and its preparation method in Sections 3 and 4.1. We then elucidate the details of the proposed deep learning (DL) architecture in Section Section 4.2. The obtained results are investigated in Section 5, followed by concluding remarks in Section 6.

\section{Related work}

    The previous works on fire detection are mainly based on image-based techniques such as classification, object detection, and semantic segmentation of visible fire imagery. Wonjae Lee et al. proposed a wildfire detection system that classifies the presence of fire in images and evaluated the performance of AlexNet, GoogLeNet, and VGG, along with their modified variants \cite{7889305}. Zhentian et al. have trained YOLOV3 for fire detection and reported a recognition rate of 91\% \cite{9163816}. An ensemble-based object detection method using YOLOV5 EfficientDet is proposed in \cite{xu2021forest} for detection and EfficientNet to capture global information about the fire. Their study showed a decrease in false positive rate by 51.3\% on 
    public datasets. Yo Zhao et al. have proposed a deep learning architecture, called Fire-Net by stacking convolutional and pooling layers for fast localization and segmentation of fire in aerial images with an accuracy of 98 \% on standard 'UAV\textunderscore Fire' dataset \cite{zhao2018saliency}. A similar method is proposed in \cite{zhang2022forest} and \cite{wang2022real} by using a deep learning architecture to classify frames into "fire with smoke", "fire with no smoke", and "no fire" using FLAME datasets \cite{shamsoshoara2021aerial,swyw-6j78-22}. However, most of these methods are limited to images, whereas real-time data for fire detection will be mostly in terms of video feed. Further, they ignore the fact that video feeds contain temporal patterns of smoke that can help locate the origin of smoke or equivalently the fire spots. Temporal analysis of such patterns, especially in the presence of clouds and other white patterns, can be extremely useful for precise fire detection.    
    This concept is used as a key idea in our method to detect obscured fire positions.

    Some works take a slightly different approach and analyze fire image patch by patch instead of one-shot analysis of the entire image. For instance, a CNN-based deep learning method is proposed in \cite{zhang2016deep} which classifies the image first and then performs a patch-level analysis to offer more detailed information. They applied their method to video frames and performed patch-wise detection. They reported a 97 \% detection accuracy on their own dataset. Still, this method does not consider the temporal relationships for fire classification since it treats the video frame as a sequence of independent still images. In a similar work, Gwangsu Kim et al\cite{kim2019fire}. proposed an algorithm that collects features of video frames using a pre-trained VGG and stacks them together to be passed through a series of fully connected layers to classify the presence of fire in the video clips. However, this method which restricts fire classification only during a visible fire deemed inefficient in capturing obscured fire.

    Anshuman et al. \cite{dewangan2022figlib} have proposed SmokeyNet - a deep learning algorithm that offers stacking CNN, LSTM, and Vision Transformer to detect smoke in the video feed captured by stand-alone cameras. However, this method is not directly applicable to aerial imagery. 
    Some other research works take advantage of Infra-Red (IR) cameras for more accurate fire positioning. 
    For instance, Chi Yuan et al. proposed an algorithm that uses brightness and motion clues with histogram-based segmentation and optical flow to segment fire in IR images \cite{yuan2017fire}. Another example is Norsuzial et al.'s work that offers an Image processing-based approach to convert IR images to YCbCr color and use a wavelet analyzer to detect fire \cite{ya2021image}. Also, a DL architecture is proposed in \cite{9953997} that analyzes dual-feed imagery captured by side-by-side RGB and IR cameras for precise fire positioning.
    
\begin{figure*}[htbp]
    \centering
    \includegraphics[width=\textwidth, height=0.42\textwidth]{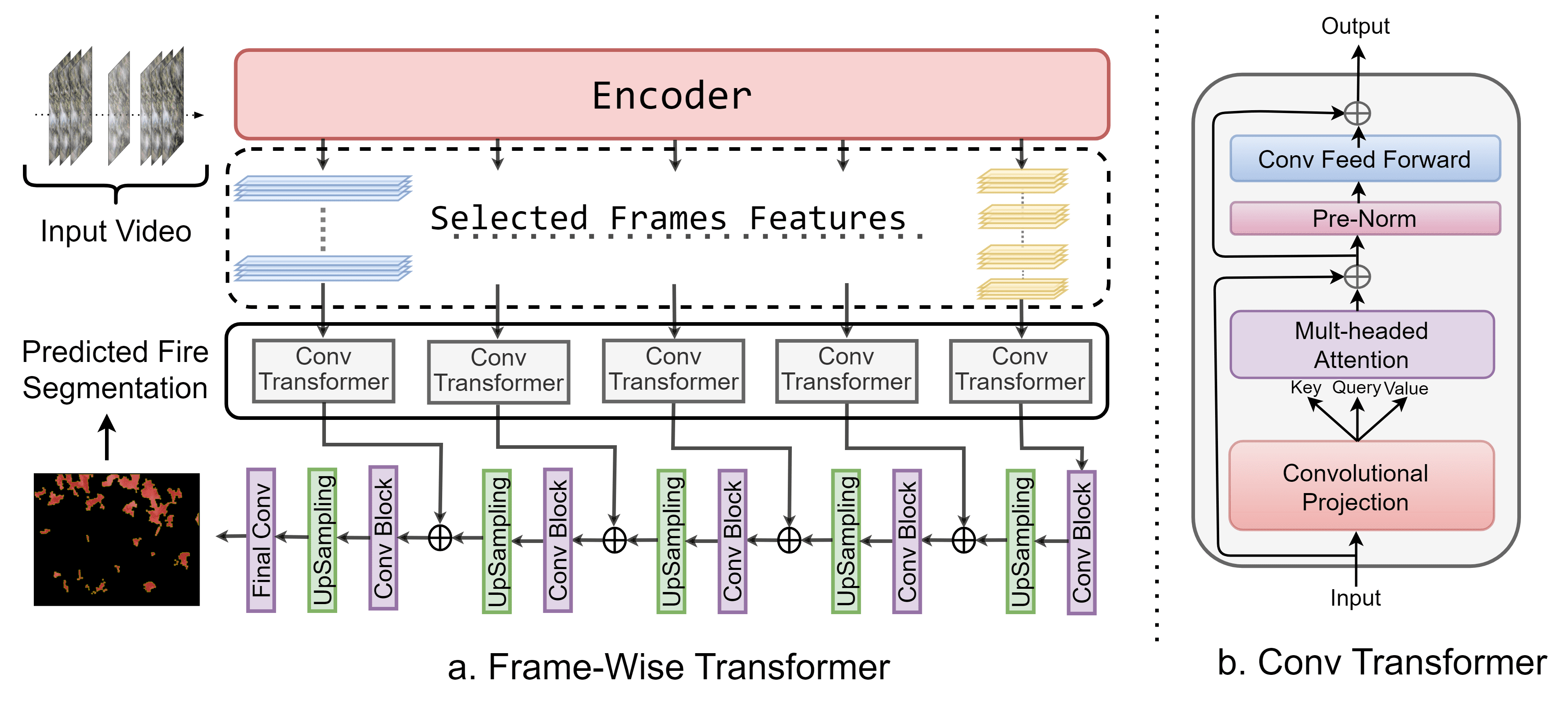}
    \caption{Proposed Deep Learning architecture - Frame Wise Transformer.} 
    \label{fig:architecture_transformer}
\end{figure*}
    
    Although these methods yield high accuracy taking advantage of thermal information captured by IR cameras, they incur an extra monitoring cost for their reliance on pricey IR cameras. Also, they are not suitable for processing existing drone-based and satellite-based datasets that include only RGB imagery. 
    In contrast to these works, our method uses only RGB videos for detecting both visible and obscured fire flames in an economical way.

\section{Data Preparation}

We investigated different fire datasets and found out that FLAME2 dataset \cite{shamsoshoara2021aerial} provides the most suitable testing condition because it has videos consisting of visible and obscured fire regions with smoke. Also, it offers a dual mode RGB/IR imagery that simplifies evaluating the proposed framework.  


FLAME2 consists of seven pairs of lengthy RGB and IR video feeds taken by seven drones equipped with side-by-side RGB/IR cameras. The videos were taken from a prescribed fire region in Kaibab National forest in Northern Arizona. For experimentation purposes, we have carefully cropped the parts of the video where the camera is relatively stationary and there is a high alignment between the RGB and IR camera viewpoints, so IR stands as a legitimate ground truth. The selected video segments are split into clips of 20 consecutive frames to train our deep neural network, where each clip is considered a training sample. In total, we have used 839 video samples for training(594) and evaluating(255) our method. These video samples consist of a diverse set of positive samples consisting of visible and obscured fire regions shot at different locations and negative samples where fire is absent, which are included to make the model robust.



\section{Method}

\subsection{Data Pre-Processing: IR to RGB Labels}

The IR images consist of heat maps corresponding to the temperature of different regions on the image. The place where the fire is present generally has a higher temperature, and the pixel values in the heat map are closer to the maximum. We have extracted the ground truth for training data where the fire is present by processing the IR image with a series of hand-crafted image processing methods. The sequence of operations performed on the IR image is presented below.

\vspace{5pt}
IR Image $\rightarrow$ Smooth Image $(5 \times 5 \text{ AvgPool})$ $\rightarrow$ Hard Thresholding $\rightarrow$ Dilation $(5 \times 5$, 2 times$)$ $\rightarrow$ Fill (flood fill) $\rightarrow$ Erosion $(5 \times 5$, 1 time$)$ $\rightarrow$ Remove small objects $(200$ pixel$)$ $\rightarrow$ Ground Truth.

\vspace{5pt}

\begin{figure}[!ht]
\begin{center}
\setlength\belowcaptionskip{-2\baselineskip}
\centerline{\includegraphics[width=1\columnwidth]{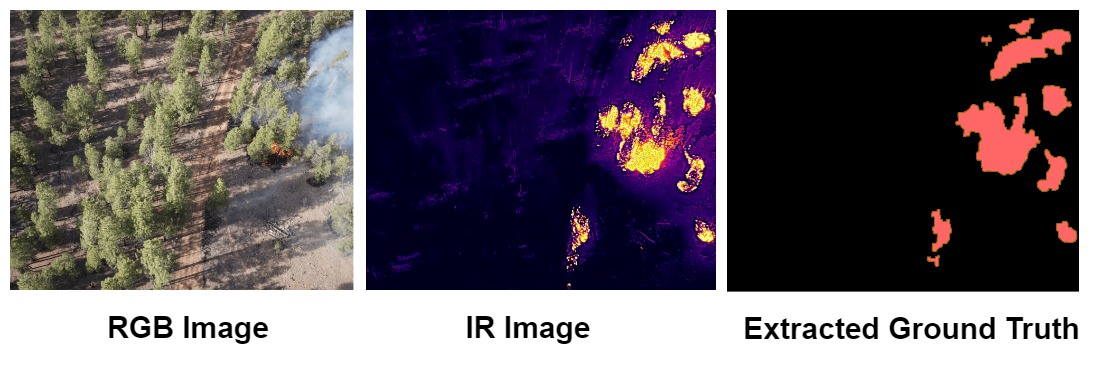}}
\caption{Obtaining Ground truth fire locations from IR images.}
\label{fig:atp}
\end{center}
\end{figure}

We initially smooth out the image and reduce the noise level using a low-pass filter, then use hard thresholding to select the regions of high-temperature values corresponding to fire. The resulting image is dilated to fill in the small spaces and make the fire boundary smooth; the spaces not filled in the previous operations are filled using flood fill, resulting in a complete blob of fire location. The image is eroded to compensate for 
the effect of dilation applied earlier. Finally, the small blobs that are likely to be representative of noise are removed. The extracted ground truth can been seen in Fig. \ref{fig:atp}

Note that we use IR images to identify ground truth fire locations when training the model, but in runtime (new monitoring tasks), we only use RGB images, so our method does not require expensive IR cameras on site.
We treat each video clip with 20 frames as a sample and use the obtained ground-truth binary fire map from the last IR frame as our reference for the entire clip.

\vspace{10pt}

\begin{figure*}[htbp]
    \centering
    \includegraphics[width=\textwidth, height=0.4\textwidth]{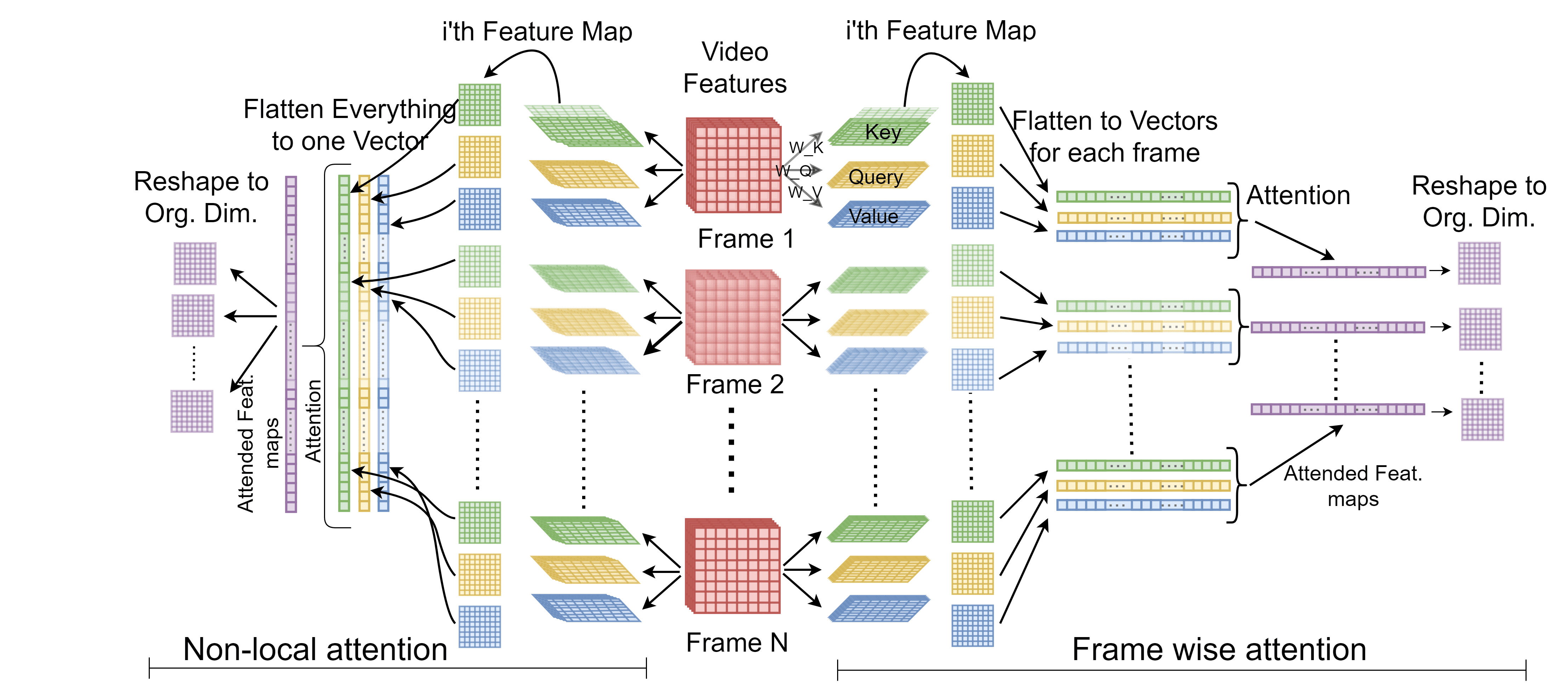}
    \caption{Comparision of Non-local vs Frame-wise attention} 
    \label{fig:attention_comparision}
\end{figure*}

\subsection{Deep Learning Architecture}

The proposed deep Learning architecture ‘Frame-wise Transformer’ comprises three stages, as shown in Fig. \ref{fig:architecture_transformer}. The first stage involves extracting features from the individual frames, which is done using a pre-trained encoder. Here, we used VGG-16 \cite{simonyan2015deep}. The next stage focuses on aggregating the sequential features of the video frames with a convolutional transformer; we have considered utilizing features at different stages of the encoder as followed in standard semantic segmentation architectures like U-net\cite{ronneberger2015unet}. Furthermore, we have examined different frame rates at different resolutions for aggregation to obtain the most efficient architecture to exploit temporal relations. The later stage consists of a decoding network with a series of convolutional and upsampling blocks that combine information at different resolutions to obtain the segmentation map.

\subsection{Convolutional Transformer}
    
We have formulated a convolutional transformer inspired by the transformer\cite{DBLP:journals/corr/VaswaniSPUJGKP17} in Natural Language Processing by treating each frame of the video as a word and feature map from the encoder as word embedding and feature maps of a series of frames as a series of word embeddings. Multiple feature maps from the encoder can be equivalent to having multiple heads. These heads can be further scaled to increase the expressiveness of the features by multiplying the input channels with $n$ heads while projecting the input into key, query, and value maps. Then, a proposed Frame-wise attention is applied to find the relations between the frames and finally fed to a convolutional block. The overall block diagram is shown in Fig. \ref{fig:architecture_transformer}(b).

We have employed residual connections and normalization around each of the sublayers, and we have chosen batch normalization instead of layer normalization, which is usually preferred in CNNs. Positional embeddings were added to the extracted video features before the convolutional projection.

\subsubsection{Convolutional Projection}
The input($x_i$) from the encoder is projected into key ($x_k$), query ($x_q$), and value ($x_v$) using a conv3d \cite{tran2015learning}, with a kernel size of 1 at the time axis, which is equivalent to convolving feature maps
only in spatial dimension as we want to separate 
information between space and time, where the correlations between the frames are handled by attention in subsequent operations. The projection is shown in Eq. (\ref{eq:convolutional_projection}) to have convolution only in the spatial dimension of the features and time dimension will be handled by the subsequent attention operation. 



\begin{align}
\label{eq:convolutional_projection}
x_i &\rightarrow \text{conv3d}(1 \times k \times k, \text{in\_c}, \text{out\_c})_{(k, q, v)} \rightarrow x_k, x_q, x_v,
\end{align}
where $k$ is the kernel size, $\text{in\_c}$ is the number of input channels, and $\text{out\_c} = n\_heads \times \text{in\_c}$ is the output size.



\subsubsection{Self Attention}

The scaled dot product provides attention to features across the frames. Initially, we tried the attention computation using the non-local block published in \cite{DBLP:journals/corr/abs-1711-07971}, which models space-time attention by flattening all the features into one single vector and performing the dot product and subsequent matrix multiplication,
 as shown in Eq. (\ref{eq:non-local-attention}). 
\begin{align}
\label{eq:non-local-attention}
\nonumber
\text{Non-local Attention}(Q, K, V) &= \\
&\hspace{-10em}\text{softmax}\left(\frac{Q_{(C \times THW)} \cdot K_{(C \times THW)}^T}{\sqrt{d_k}}\right) \cdot V_{(C \times THW)}
\end{align}
where $C$ is the number of input channels, $T$ is the sequence length (number of consecutive frames), and $H,W$ is the dimension (height, width) of the feature space.

However, this method mixes up the information in space and time causing ambiguity in distinguishing information between space and time. In the NLP universe, one can imagine this operation as finding attention within the word embedding while finding attention with other words. 
Therefore, we have modified the computation by providing attention between the frames but not within the frame, which finds relations between the frames while separating within the frame information. 
The proposed attention mechanism and its distinction with non-local attention are presented in Fig. \ref{fig:attention_comparision}.
The modified attention mechanism can be expressed as 
\begin{align}
\label{eq: Frame-Wise attention}
\nonumber
\text{Frame-Wise Attention}(Q, K, V) &= \\
&\hspace{-13em}\text{softmax}\left(\frac{Q_{(C \times T \times HW)} \cdot K_{(C \times T \times HW)}^T}{\sqrt{d_k}}\right) \cdot V_{(C \times T \times HW)}
\end{align}

\begin{figure*}[htbp]
    \centering
    \includegraphics[width=\textwidth]{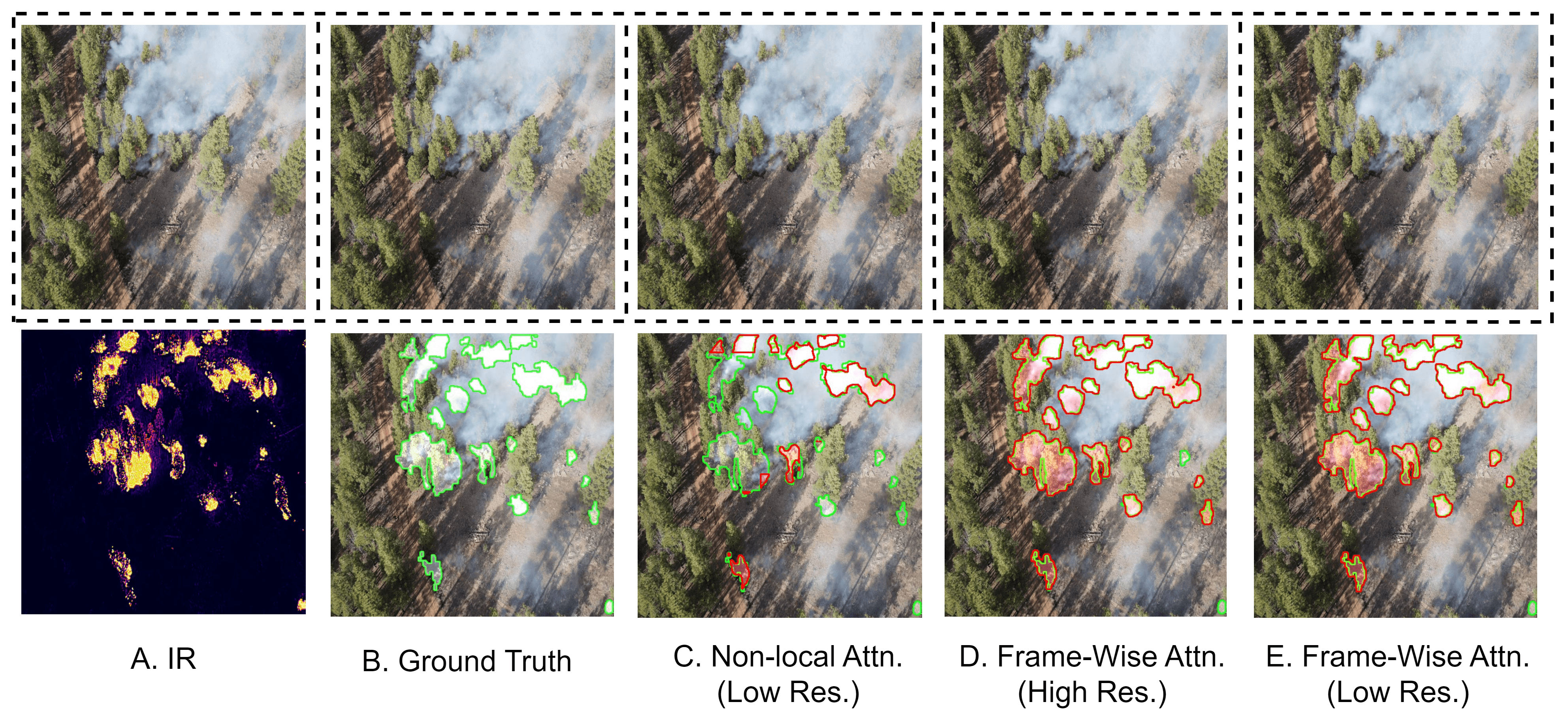}
    \caption{Top row is a few frames of sample video. Bottom row, A is the IR Image from where ground truth is extracted. The Green Regions correspond to Ground Truth and the Red are predictions.}
    \label{fig:qual_res}
\end{figure*}

\subsubsection{No-feature Concatenation}

Post the attention operation, unlike the NLP transformer, where the output of multiple heads is concatenated and averaged out, we have used a convolutional block to reduce the number of multi-headed self-attention feature maps to the original input number.

\subsubsection{Convolutional Feed Forward}

Similar to using a fully connected feed-forward network in the transformer, we have employed a convolutional feed-forward network, with a series of convolution, batch norm, and relu operations.

\vspace*{-\baselineskip}

\begin{align}
\label{eq: Convolutional Feed Forward}
    \nonumber
    &\text{conv\_feed\_forward} := \text{conv3d}(1 \times k \times k, \text{in\_c}, \text{hidd\_c}) \rightarrow \\
    &~~~~~~\text{ReLU}() \rightarrow \text{BatchNorm}(\text{hidd\_c}) \rightarrow \nonumber \\
    &~~~~~~\text{conv3d}(1 \times k \times k, \text{hidd\_c}, \text{in\_c}) \rightarrow \text{ReLU}()
\end{align}
where $k$ is the kernel size, $in\_c$ is the number of input channels, and $hidd\_c$ is the number of hidden channels


\subsection{Frame Rate Variation}


We have developed a technique for varying frame rates based on principles of natural eyesight physics. Essentially, when objects are far away, they appear smaller, and the entire scene is visible, whereas when objects are closer, only parts of the scene are visible. Leveraging this concept, we have explored two theories. The first is to use all the frame features of low resolution from the encoder and lower the use of the number of frame features in the previous higher resolutions at a chosen frame rate. The second theory inverts the first theory by using more frames at higher resolution and fewer frames at lower resolution. We got the best results using the first theory, which aligns with our eyesight-based principle, with a significant margin in computational gain compared to the second theory. The results are discussed in Section 5.



\subsection{Loss Function}

We have used Dice loss \cite{Sudre_2017} to measure the alignment between the ground truth and the detected fire regions by the architecture to train the network. The Dice coefficient measures the alignment (similarity) 
between two corresponding segments by computing the overlap coefficient. The overlap coefficient ranges from 0 to 1, with 1 indicating a perfect match between the two segments. The Dice loss function is defined as one minus the Dice coefficient, with the objective of minimizing the loss during training. More specifically, we have

\begin{align}
\label{eq:dice}
\nonumber 
DiceLoss &= 1- \frac{2 \times \text{Intersection}}{\text{Predicted} + \text{Ground Truth}} \\
&=1 - \frac{2 \sum_{i=1}^{n}p_i g_i + \epsilon}{\sum_{i=1}^{n}p_i^2 + \sum_{i=1}^{n}g_i^2 + \epsilon},
\end{align}
where $p_i$ and $g_i$ are the predicted and ground truth segmentation masks, respectively, for the $i$-th pixel in the image. The summations are taken over all $n$ pixels in the image. The $\epsilon$ term is a small constant added to the denominator to prevent division by zero.

\subsection{Inference}

\begin{figure}[!ht]
\begin{center}
\setlength\belowcaptionskip{-2\baselineskip}
\centerline{\includegraphics[width=0.95\columnwidth]{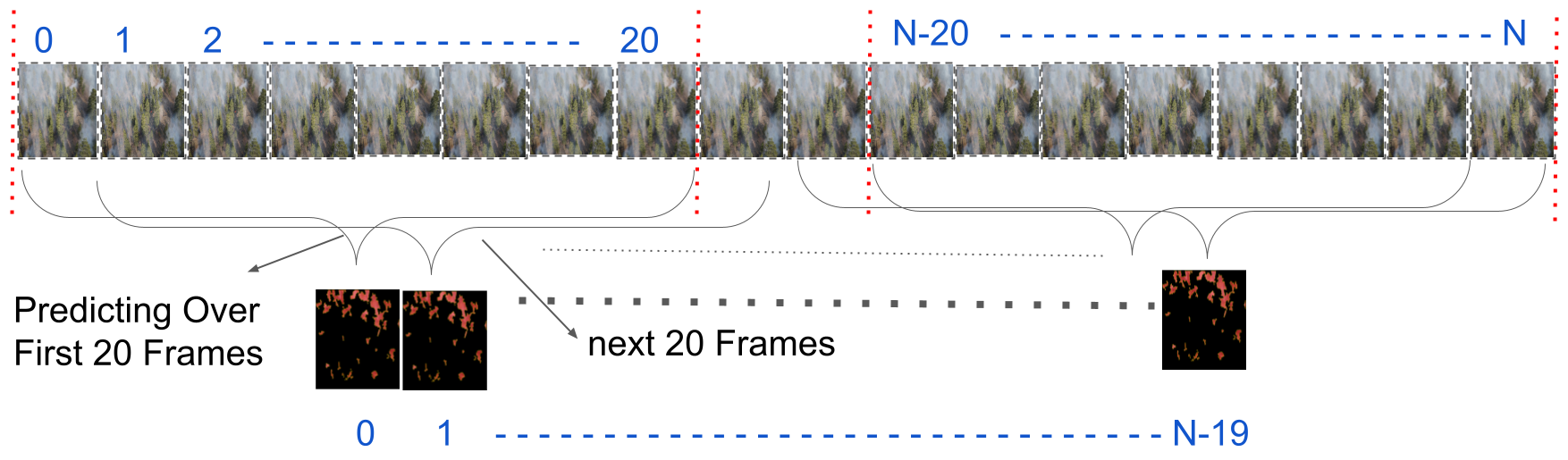}}
\caption{Inference method: each output map results from sequential processing of 20 preceding input frames.}
\label{fig:inference}
\end{center}
\end{figure}

The inference in real-time, where videos are lengthy, is made by taking frames at a window size of 20 and sliding the window over the video at a stride of one as shown in Fig. \ref{fig:inference}.

\vspace{2pt}
\begin{table*}[t]
  \centering
  \begin{tabular}{|c|c|c|c|c|c|c|c|c|c|}
    \hline
    \multirow{2}{*}{\centering Threshold} 
    & \multicolumn{2}{c|}{\centering Speed} & \multicolumn{3}{c|}{F1 Score} & \multicolumn{3}{c|}{FG Dice}& \multicolumn{1}{c|}{CL-Accuracy}\\ 
    \cline{2-10}
        & \multicolumn{1}{|p{1.4cm}|}{\centering Frame-Wise \\ High Res. }
        & \multicolumn{1}{|p{1.4cm}|}{\centering Frame-Wise \\ Low Res. }
        & \multicolumn{1}{|p{1.4cm}|}{\centering Non-Local \\ Low Res. }
        & \multicolumn{1}{|p{1.4cm}|}{\centering Frame-Wise \\ High Res. }
        & \multicolumn{1}{|p{1.4cm}|}{\centering Frame-Wise \\ Low Res. }
        & \multicolumn{1}{|p{1.4cm}|}{\centering Non-Local \\ Low Res. }
        & \multicolumn{1}{|p{1.4cm}|}{\centering Frame-Wise \\ High Res. }
        & \multicolumn{1}{|p{1.4cm}|}{\centering Frame-Wise \\ Low Res. } 
        & \multicolumn{1}{|p{1.4cm}|}{\centering Frame-Wise \\ Low Res.} \\
     
    \hline
    0.1 & 0.48 X & 0.27 X & 83.30 & 90.37 & 91.50 & 90.18  & 92.55 & 92.56 & 98.81\\
    0.2 & 0.50 X & 0.28 X & 86.06 & 91.66 & 92.40 & 90.82  & 92.29 & 92.69 & 97.62\\
    0.3 & 0.53 X & 0.28 X & 85.98 & 90.87 & 92.46 & 90.53  & 92.66 & 92.62 & 98.02\\
    0.4 & 0.58 X & 0.32 X & 84.22 & 89.91 & 91.07 & 90.61  & 92.46 & 92.65 & 98.02\\
    0.5 & 0.62 X & 0.35 X & 82.97 & 90.82 & 91.39 & 90.53  & 92.45 & 92.78 & 98.41\\
    0.6 & 0.69 X & 0.43 X & 80.85 & 90.58 & 91.31 & 88.39  & 92.80 & 92.76 & 97.63\\
    0.7 & 0.75 X & 0.51 X & 81.83 & 89.85 & 90.07 & 89.21  & 92.73 & 92.79 & 96.84\\
    0.8 & 0.83 X & 0.65 X & 81.92 & 88.34 & 89.24 & 90.20  & 92.62 & 92.64 & 97.23\\
    0.9 & 0.92 X & 0.82 X & 81.10 & 88.52 & 88.91 & 89.32  & 92.37 & 92.00 & 99.60\\
    1.0 & 1.00 X & 1.00 X & 79.63 & 90.87 & 90.87 & 87.72  & 92.15 & 92.15 & 98.81\\
    \hline
  \end{tabular}
  \caption{Comparision of Speed and Performance between taking more frames at high resolution (High Res.) Vs. More frames at low res (Low Res.). And Performance comparison between Non-local and Proposed Frame-Wise attention. All experiments are done using a 3x3 convolutional kernel size. X represents the number of flops performed per sample. CL-Accuracy - Clip Level classification accuracy.}
  \label{tab:High Res Vs Low Res}
\end{table*}

\section{Experiments}

In this section, we present the simulation results using the FLAME2 dataset, which consists of RGB and IR images; we have trained with pre-processed videos as explained in the Data Preparation step. We have used 594 videos for training and 255 videos for testing. Out of these, 354 of training and 155 of testing are positive examples, consisting of visible and obscure fire regions, and the rest are negative videos with no fire locations. 
All the training and testing videos are independent and include no overlapping frames.

\subsection{Training}

Our model is implemented in PyTorch lightning \cite{ronneberger2015unet} using a single node Linux machine with 2 Tesla A-100 40 GB GPUs.
The models were tuned for the best hyperparameters. We used a step-learning rate with an initial learning rate of 2e-4 with Adam Optimizer \cite{kingma2014adam}. The models were trained for 150 epochs with early stopping and a batch size of 2.

\subsection{Evaluation Metrics}

We are using a set of metrics to evaluate the quality of fire detection. We used the Fore Ground Dice score (presented in Eq. (\ref{eq:dice})) to assess the alignment quality between ground truth and the detected fire region. This assessment is particularly important when part of the fire is obscured by thick smoke to demonstrate how capable our model is in detecting such fire regions.

Another metric we used is blob-wise object detection F1 score Eq. (\ref{eq:F1 score}), which is the harmonic mean of precision and recall Eq. (\ref{eq:precision_recall}) to ensure the correctness of our predictions; it is calculated by taking each blob in the ground truth and prediction, and if the intersection area is greater than 50\%, considering it as a True Positive, else a False Positive. 

\begin{align}
\label{eq:precision_recall}
\text{Precision} &= \frac{\text{TP}}{\text{TP} + \text{FP}}, \quad
\text{Recall} = \frac{\text{TP}}{\text{TP} + \text{FN}}
\end{align}

\begin{equation}
\label{eq:F1 score}
\text{F1 Score} = \frac{2 \times \text{Precision} \times \text{Recall}}{\text{Precision} + \text{Recall}}
\end{equation}
where $TP$ is True Positive, $FP$ is False Positive and $FN$ is False Negative.




Also, we calculate clip-level classification accuracy to obtain clip-level labels, Fire and No-Fire. 
This is calculated by counting the number of fire spots in the video and comparing it with the prediction; 
if more than 50\% of spots are predicted, we will classify the video as Fire; otherwise, No-Fire. 

\begin{table}[t]
    \centering
    \begin{tabular}{|c|c|c|}
        \hline
        \multicolumn{1}{|c|}{Threshold} 
        & \multicolumn{1}{|m{2cm}|}{\centering F1-Score \\ Non-Local }
        & \multicolumn{1}{|m{2cm}|}{\centering F1-Score \\ Frame-Wise } \\
        \hline
        0.1 & 85.57 & 88.16 \\
        0.3 & 85.26 & 90.26 \\
        0.5 & 86.30 & 90.96 \\
        0.7 & 83.37 & 89.21 \\
        0.9 & 80.67 & 84.72 \\
        \hline
    \end{tabular}
    \caption{Non-local Vs. Frame-Wise using 1x1 kernel}
    \label{tab:Non-local Vs. Frame-Wise using 1x1 kernel}
\end{table}

\subsection{Quantitative Results}

We have experimented with our architecture from two perspectives. First, we investigate the architecture with the Non-Local attention method versus the proposed Frame-Wise attention method. Next, we study the impact  of using various depths at different resolutions by comparing 
High Res. versus Low Res. at different thresholds. 

\subsubsection{Thresholds}

The thresholds are the values that indicate the percentage of frames used in the subsequent resolution starting with a use of 100\% frames. For example, in Low Res., if the value is $\alpha=0.9$, at resolution (H x W)/16 which is the lowest, 100\% of the frames will be used, and in the subsequent resolution (HxW)/8, 90\% of the frames (the most recent ones) will be used and again in the next resolution 90\% of the remaining frames in the previous resolution will be used ($\alpha^2=0.9\times0.9$) and so on. Therefore, the number of frames at each layer follows exponential decay, layer by layer.
Likewise, for High Res. 100\% of the frames will be used at the highest resolution, (HxW)/2, and the number of frames decreases, with the rate of $\alpha=0.9$, for the lower resolutions. 



\subsubsection{Non-Local Vs. Frame-Wise Attention.}

\begin{figure}[!ht]
\begin{center}
\setlength\belowcaptionskip{-2\baselineskip}
\centerline{\includegraphics[width=1\columnwidth]{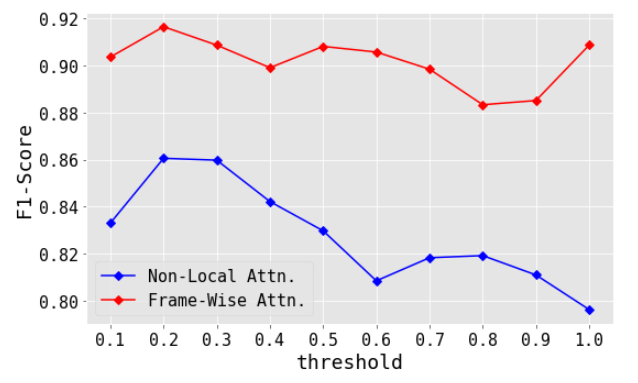}} 
\caption{F1 Score Non-Local Vs Proposed Frame-Wise Method}
\label{fig:non-local vs frame-wise}
\end{center}
\end{figure}

We have presented the F1 score and Fore Ground Dice coefficients for both methods in Table \ref{tab:High Res Vs Low Res}. For the results in this table, we have used a 3x3 kernel for convolution in the conv-transformer. The proposed method has outperformed the non-local method.  The best F1-Score using Frame-Wise at the threshold 0.3 has more than 7\% margin than using Non-Local method The F1-Score graph in Fig. \ref{fig:non-local vs frame-wise} clearly shows the difference.

\begin{figure}[!ht]
\begin{center}
\setlength\belowcaptionskip{-2\baselineskip}
\centerline{\includegraphics[width=1\columnwidth]{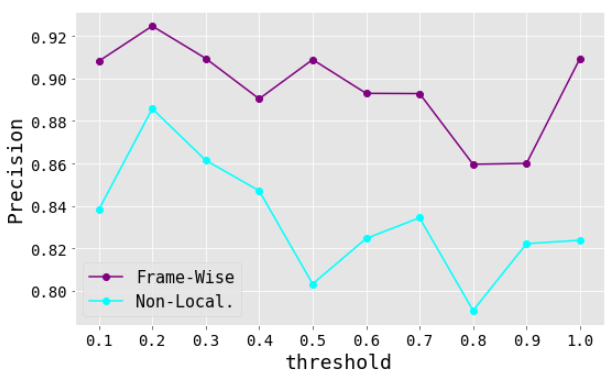}} 
\caption{Precision - Non-Local Vs Proposed Frame-Wise Method}
\label{fig:precision}
\end{center}
\end{figure}

\begin{figure}[!ht]
\begin{center}
\setlength\belowcaptionskip{-2\baselineskip}
\centerline{\includegraphics[width=1\columnwidth]{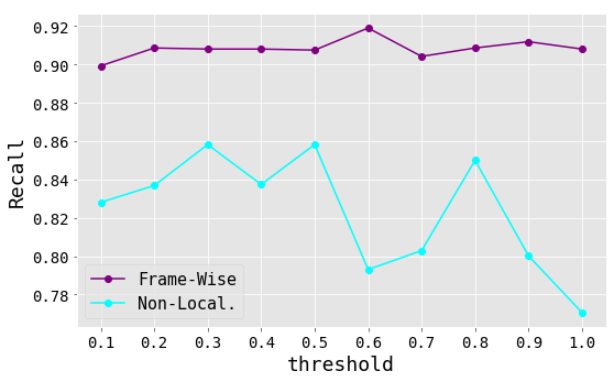}} 
\caption{Recall - Non-Local Vs Proposed Frame-Wise Method}
\label{fig:recall}
\end{center}
\end{figure}

We have also experimented with a 1x1 kernel with few thresholds to make an exact match with the kernel size used in the non-local block original paper. Similar to the 3x3 kernel, the Frame-Wise method performed better with a margin of around 5 \% at the 0.3 threshold.



\subsubsection{High Res. Vs. Low-Res and Computational Gain}

\begin{figure}[!ht]
\begin{center}
\setlength\belowcaptionskip{-2\baselineskip}
\centerline{\includegraphics[width=1\columnwidth]{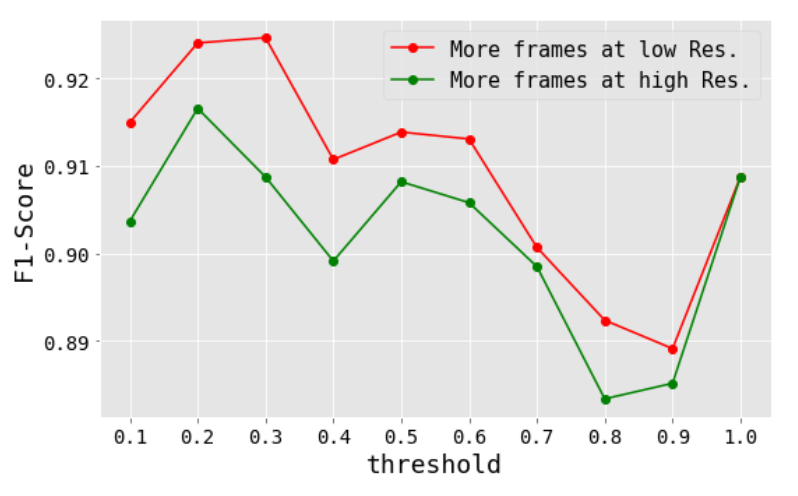}}
\caption{High Res. Vs Low Res.}
\label{fig:frame_rate}
\end{center}
\end{figure}

We also have displayed results of using more frames at low resolutions vs. using more frames at high resolution in Table \ref{tab:High Res Vs Low Res} and Fig. \ref{fig:frame_rate}. It can be seen that using more frames at low resolution leads to better results, which 
verifies the logic behind our justification 
in Section 4. Also, the best results were obtained at a threshold value of $\alpha=0.3$ for the Frame-Wise attention, where the computational cost is reduced by 72 times 
compared to using all frames. 
From the speed column in Table \ref{tab:High Res Vs Low Res}, we can infer that Low Res. takes fewer flops compared to High Res. at any threshold.

\subsection{Qualitative Results}

The qualitative results are displayed in Fig. \ref{fig:qual_res}. The top row shows a few frames of a sample input video. In the bottom row, A is the IR Image image of a sample frame, and B is the ground truth extracted from it and displayed over the image. Image C is the result using non-local attention, which fails in detecting many blobs. D and E are the results of the proposed frame-wise attention method for High Res. (more frames at higher resolution) and Low Res (more frames at lower resolution) approaches. The High Res. misses some blobs compared to the best output for Low Res.


\section{Conclusion}

This work is devoted to detecting obscured fire flames by analyzing video feeds captured by drones. All existing works, to the best of our knowledge, solve fire detection through image-based segmentation and object-detection methods. This neglects the fact that temporal analysis of video frames can recover motion patterns of smoke that can be indicative of obscured fire flames. Therefore, existing work fails to recover fire flames obscured by smoke patterns. To address this matter, we developed a novel architecture based on the frame-wise convolutional transformer that uses between-frame attention to capture relationships between frames to help locate obscured fire flames. 
We investigated the trade-off between feature resolution and temporal depth and observed that it is advantageous to use a higher depth for lower resolution, and lower depth for higher resolution. This reflects the fact that long-term motions can be captured in low-resolution feature spaces. Applying our model to FLAME2 datasets resulted in almost full success in clip-level fire classification while achieving about Dice score of 92.6\%, and precision and recall scores of 93.2\% and 91.73\% in detecting fire flames. Our ablation study shows that our proposed attention mechanism outperformed the well-known non-local block with more than 7\% improvement in F1 score.

We examined the interplay between spatial resolution and temporal depth by experimenting with various frame rates and resolutions. Our findings indicate that employing more frames at lower resolutions offers enhanced performance and computational efficiency compared to uniform depth allocation across resolutions. This versatile framework can be tailored for other video-based deep-learning tasks with slight modifications to meet specific task requirements.

To support our research, we curated a dataset from the FLAME2 dataset, carefully selecting video frames where the drone is stationary and synchronized with IR images to mitigate alignment discrepancies. Ground truth labels were generated using corresponding IR videos and image-processing operations. Through meticulous visual verification, we ensured frame-level synchronization between RGB and processed IR videos.

In conclusion, our methodology presents a robust solution for obscured fire detection using video analysis. The 'Frame-Wise Transformer' architecture, informed by our tailored attention mechanisms and frame aggregation strategies, showcases its potential for video-based deep learning applications.


\end{document}